# Improving Clinical Documentation with AI: A Comparative Study of Sporo AI Scribe and GPT-4o mini

Chanseo Lee[1,2], Sonu Kumar,[1] Kimon A. Vogt,[1] Sam Meraj[1]

1. Sporo Health, Boston, MA
2. Yale School of Medicine, New Haven, CT

**Abstract**

AI-powered medical scribes have emerged as a promising solution to alleviate the documentation burden in healthcare. Ambient AI scribes provide real-time transcription and automated data entry into Electronic Health Records (EHRs), with the potential to improve efficiency, reduce costs, and enhance scalability. Despite early success, the accuracy of AI scribes remains critical, as errors can lead to significant clinical consequences. Additionally, AI scribes face challenges in handling the complexity and variability of medical language and ensuring the privacy of sensitive patient data. This study aims to evaluate Sporo Health's AI Scribe, a multi-agent system leveraging fine-tuned medical LLMs, by comparing its performance with OpenAI's GPT-4o mini on multiple performance metrics. Using a dataset of de-identified patient conversation transcripts, AI-generated summaries were compared to clinician-generated notes (the ground truth) based on clinical content recall, precision, and F1 scores. Evaluations were further supplemented by clinician satisfaction assessments using a modified Physician Documentation Quality Instrument revision 9 (PDQI-9), rated by both a medical student and a physician. The results show that Sporo AI consistently outperformed GPT-4o mini, achieving higher recall, precision, and overall F1 scores. Moreover, the AI-generated summaries provided by Sporo were rated more favorably in terms of accuracy, comprehensiveness, and relevance, with fewer hallucinations. These findings demonstrate that Sporo AI Scribe is an effective and reliable tool for clinical documentation, enhancing clinician workflows while maintaining high standards of privacy and security.

**Introduction**

In recent years, the advent of AI-powered medical scribes has garnered significant attention as a potential solution to the documentation burden faced by healthcare providers. AI scribes, such as those powered by natural language processing (NLP) algorithms, offer real-time transcription and automated data entry into EHRs. These systems aim to further streamline administrative tasks by eliminating the need for human scribes, thereby reducing costs and increasing scalability. AI scribes have already shown promising results in early trials, and can also be integrated with additional clinical decision support AI, alerting physicians to potential errors or missed diagnoses, which adds a layer of safety to the documentation process.[1][2]

However, the success of scribes depends heavily on the accuracy of the documentation they produce. Accurate transcription is critical in ensuring that patient records are reliable, up-to-date, and reflective of the care provided, as even small errors can lead to significant clinical consequences. Furthermore, accuracy of information is necessary but not sufficient, as inclusion of not just correct but also all of the salient details from the patient-physician conversation is crucial. That being said, maintaining accuracy for AI scribes poses unique challenges due to the variability of medical language, the complexity of clinical interactions, and the need to understand contextual nuances.[3] While human scribes can often clarify or correct errors based on clinical knowledge, AI scribes may struggle with ambiguous terms or less structured conversations, potentially introducing discrepancies into the medical record. Additionally, ensuring patient privacy and security is another concern for AI-powered solutions, as the handling of sensitive data must comply with regulations like HIPAA, and any mismanagement of this information could pose ethical and legal risks.[1] The importance of accuracy in scribe-generated documentation—whether by human or AI—cannot be overstated, as it directly impacts patient safety, treatment decisions, and overall healthcare quality.

At Sporo Health, we are leading the way in developing fine-tuned AI models designed specifically for use in AI scribing. Our innovative models address many of the key challenges identified with AI scribes, including issues of accuracy, usability, and contextual understanding. With a focus on creating solutions that are not only precise but also ensure comprehensive capture of all relevant details, our AI scribes are privacy-compliant and resistant to hallucinations, providing a safer, more reliable tool for clinical documentation. In this case study, we will present the first in a series of evaluations of our AI models, beginning with our Sporo AI Scribe against OpenAI GPT-4o mini, as we aim to demonstrate its effectiveness in solving these critical challenges in AI scribing.

**Methods**

We collected a dataset of de-identified patient conversation transcripts from one of our partner clinics, generated using Sporo Health's proprietary speaker labeling and transcription AI. For each patient encounter, the piloting clinician provided Subjective, Objective, Assessment, and Plan (SOAP) notes through the scribe platform, which we designated as the ground truth. Sporo Health's AI agentic workflow, along with GPT-4o mini hosted within Azure Playground, were then tasked using zero-shot prompting with generating SOAP-formatted summaries from the same transcripts. These AI-generated summaries were subsequently compared to the clinician's ground truth notes using various quantitative evaluation metrics.

**Clinical content recall (sensitivity)** was defined as the proportion of relevant clinical information from the clinician's ground truth summary that was accurately captured in the AI-generated summaries. To evaluate recall, salient clinical items were manually extracted from each conversation into an inventory. Recall was then calculated by dividing the number of correctly included items from the inventory by the total number of relevant items identified in the inventory.

**Clinical content precision (positive predictive value)** was defined as the proportion of information in the AI-generated summary that was both accurate and relevant when compared to the clinician's ground truth. Precision was calculated by dividing the number of correctly included items in the AI-generated summary by the total number of items in the AI summary, including any additional or incorrect items. This metric reflects the accuracy and relevance of the AI-generated content without introducing extraneous or inaccurate details.

The **F1 score** is used as a balanced metric to combine both clinical content precision and recall, providing a single measure of the AI-generated summaries' performance.[4] It represents the harmonic mean of precision and recall, ensuring that both the accuracy of relevant information captured (precision) and the completeness of that information (recall) are taken into account. The F1 score was calculated using the formula:

$$F1 = 2 \times \frac{\text{Precision} \times \text{Recall}}{\text{Precision} + \text{Recall}}$$

This metric is particularly useful for evaluating the overall effectiveness of the AI model when there is a need to balance precision and recall in the generated summaries.

In addition to the objective accuracy metrics, clinical user satisfaction was evaluated in conjunction with accuracy using the **modified Physician Documentation Quality Instrument revision 9 (PDQI-9)**. The original PDQI-9 employs a 5-point Likert scale across nine attributes to assess the quality of clinical notes. This was then modified by Tierney et al. into a ten-item inventory to better fit the metrics relevant to ambient AI documentation, and is widely used to evaluate AI-generated clinical notes.[5],[6] The attributes evaluated in the modified PDQI-9 are detailed in **Table 1.** Prior to evaluation, the physician-generated summary and the two AI-generated summaries were sanitized of any HIPAA-protected information.[7] A certified physician and a medical student first reviewed the physician-generated summary, which served as the ground truth. They then reviewed the two AI-generated summaries—blinded to their source—and assessed the quality of the AI-generated notes using the PDQI-9. The evaluation particularly focused on both the quality of the notes and their alignment with the style, content, and utility of the physician-generated notes.

| PDQI-9 Attribute | Explanation |
|---|---|
| Accurate | The note does not present incorrect information. |
| Thorough | The information presented is comprehensive and lacks omissions. It contains all information that is relevant to the patient. |
| Useful | The information presented is relevant and provides valuable information for patient management. |
| Organized | The note is formatted in a way that is coherent and easy to comprehend. It helps the reader to understand the patient's story and the management of their clinical case. |
| Comprehensible | The note is straightforward, with no unclear or hard-to-understand sections. |
| Succinct | The note does not contain redundant information and presents relevant information in a concise, direct manner. |
| Synthesized | The note demonstrates the AI's comprehension of the patient's condition and its capability to formulate a care plan. |
| Internally Consistent | The facts presented within the note are consistent with each other and do not contradict the patient's story, each other, or known medical knowledge. |
| Free from Bias | The note is unbiased and includes only information that can be verified by the transcript, without being influenced by the patient's characteristics or the nature of the visit. |
| Free from Hallucinations | The information in the note aligns with the content of the transcript, without any factual inaccuracies or AI-generated hallucinations. |

**Table 1.** Items of the modified PDQI-9 for AI-generated summary evaluation.

**Results**

Twenty salient clinical items were collected from the clinician-generated summary, which included items relevant to the chief complaint (ex. "XX presents today for a follow-up appointment regarding their chronic headaches and to discuss potential treatment options, including Botox"), social history (ex. career-related items such as "XX is studying for their real estate license"), external emotions such as stress and anxiety, portions of the physical exam, and follow-up items such as scheduling an MRI. Then, the number of items were counted within each of the summaries, and compiled into the respective evaluation metric presented in **Table 2.**

| Metric | Sporo Summary Score | GPT-4o mini Summary Score |
| --- | --- | --- |
| Clinical Content Recall | 75% | 60% |
| Clinical Content Precision | 83% | 75% |
| F1 Score | 79% | 67% |

**Table 2.** Objective quantitative evaluation metrics for AI-generated summaries.

The healthcare-specific Sporo AI demonstrated superior accuracy and comprehensiveness across all evaluated metrics. While both AI-generated summaries omitted some items, Sporo AI consistently delivered reliable results despite occasional gaps. Clinician satisfaction, measured using the PDQI-9, is detailed in **Table 3**. The inventory revealed that there was a slight difference in overall satisfaction favoring Sporo's summary over GPT-4o mini's summary. A notable finding during the evaluation was that both evaluators identified additional clinical items in the AI-generated notes that were not present in the physician-generated note. This added level of comprehensiveness aligns with recent literature suggesting that AI-powered clinical documentation can, in certain aspects, surpass human clinicians in capturing relevant information.[8]

|  | Sporo Summary | | GPT-4o mini | |
|---|---|---|---|---|
| **PQDI-9 Attribute** | *Evaluator 1* (Medical Student) | *Evaluator 2* (Physician) | *Evaluator 1* (Medical Student) | *Evaluator 2* (Physician) |
| Accurate | 5 | 4 | 5 | 4 |
| Thorough | 4 | 5 | 3 | 5 |
| Useful | 5 | 5 | 3 | 5 |
| Organized | 4 | 4 | 3 | 4 |
| Comprehensible | 5 | 5 | 5 | 5 |
| Succinct | 4 | 5 | 5 | 5 |
| Synthesized | 4 | 4 | 3 | 3 |
| Internally Consistent | 5 | 5 | 5 | 5 |
| Free from Bias | 5 | 5 | 5 | 5 |
| Free from Hallucinations | 4 | 5 | 4 | 4 |
| **Total** | 45/50 | 47/50 | 41/50 | 45/50 |
| **Average** | **46/50** | | **43/50** | |

**Table 3.** Modified PDQI-9 evaluation of AI-generated summaries.

**Discussion**

In summary, the evaluation of Sporo AI Scribe in comparison to GPT-4o mini demonstrated the clear superiority of Sporo AI in both objective accuracy metrics and clinician-centered satisfaction. Sporo AI consistently outperformed GPT-4o mini in recall, precision, and overall F1 score, capturing more relevant clinical information while reducing errors and omissions. Clinician satisfaction, as assessed through the modified PDQI-9, further emphasized Sporo AI's advantages. The AI scribe generated notes that were more comprehensive and accurate, closely aligning with the clinician's ground truth while maintaining clarity and relevance with minimal hallucinations. Additionally, Sporo AI is fully HIPAA-compliant, adhering to the highest standards of privacy and security, and it prioritizes bias-free outputs, ensuring reliable and impartial clinical documentation. Furthermore, Sporo AI included accurate ICD-10 codes in the notes, enhancing documentation and reducing billing inaccuracies. Notably, the physician-generated note failed to capture some relevant information from the transcript, which was accurately included by the AI scribe, underscoring the AI's superior performance. These findings suggest that Sporo AI Scribe is not only a more reliable tool for clinical documentation but also better equipped to enhance clinician workflows by producing high-quality notes that meet both technical and practical needs.